# Investigating the effects of exploration dynamics on stiffness perception

Mohit Singhala,[1] *Member, IEEE* and Jeremy D. Brown,[2] *Member, IEEE*

*Abstract*— The utility of Human-in-the-loop telerobotic systems (HiLTS) is driven in part by the quality of feedback it can provide to the operator. While the dynamic interaction between robot and environment can often be sensed or modeled, the dynamic coupling of the human-robot interface is often overlooked. Enabling dexterous manipulation through HiLTS however, will require careful consideration of human haptic perception as it relates to the human's changing limb impedance at the human-robot interface. In this manuscript, we present results from a stiffness perception task run on a simple 1-DoF rotational kinesthetic device at three different angular velocities, based on participant's natural exploration strategy. We evaluated performance effects of exploration velocity as a proxy measurement for limb impedance and the results indicate the need to further investigate how the human body incorporates its knowledge of the body dynamics in kinesthetic perception under active exploration.

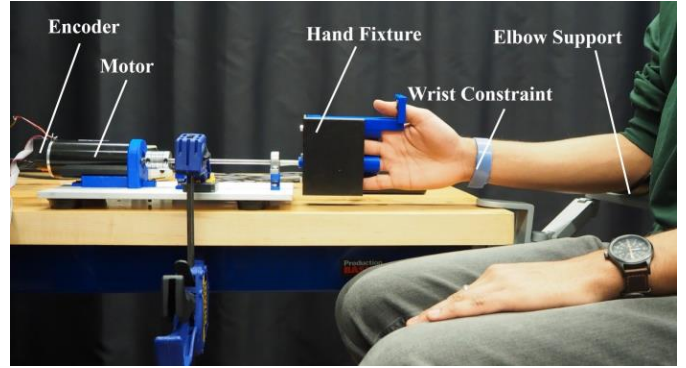

Fig. 1. Experimental setup with a motor, encoder, custom hand fixture, hook and loop wrist constraint, and the elbow support.

## I. INTRODUCTION

The human body is capable of performing a wide variety of complex manipulation tasks requiring high dexterity, often with objects of different shapes, sizes, and mechanical compliance. Performance in tasks which require integration of force and motion information (environmental impedance) [1], can be affected by how humans integrate this information into the percepts needed for task execution [2]. Particularly in the case of haptic interactions with the environment, forces exerted by the body have been shown to be regulated in terms of limb impedance [3].

It has also been shown that human haptic perception is informed by limb impedance [2], which in turn is affected by the dynamics of human exploration. Limb impedance can change significantly with the velocity of motion in upper limbs [4] and task requirements for velocity and accuracy are also known to effect muscle activation patterns [5]. Several studies have attributed the human's ability to modulate limb impedance in accordance with environmental parameters to the central nervous system (CNS) [3], [6], [7]. The literature in this field, however, is limited to either cutaneous stimuli or passive stimuli with constant physical properties. Therefore, further investigation is needed to understand the CNS's ability to account for changing body dynamics (limb impedance) while perceiving active stimuli, such as stiffness.

In this manuscript, using a single degree-of-freedom wrist-rotation kinesthetic haptic device, we investigated the effect of haptic exploration velocity on stiffness perception. In an adaptive psychophysics paradigm, Just Noticeable Difference (JND) for stiffness was evaluated at three different velocities based on each participant's natural exploration strategy. In this way, our intent was to evaluate a participant's perception in a region defined by their preferred manner of exploration.

## II. METHODS

### A. Experimental Setup

We recruited n=13 individuals (10 male, 3 female, age = 26 ±years) who were compensated at a rate of $10/hour. All participants were consented according to a protocol approved by the Johns Hopkins School of Medicine Institutional Review Board (Study# IRB00148746). The experimental apparatus consists of a custom direct drive 1-DoF rotary kinesthetic haptic device (Fig. 1). A Quanser AMPAQ-L4 linear current amplifier is used to drive a Maxon RE50 motor (200 Watt), equipped with a Maxon HEDL 5540 encoder (3 channel, 500 CPT). A Quanser QPIDe PCI data acquisition card is used for data acquisition and controlled via a MATLAB/Simulink and Quarc real-time software interface at a sample rate of 1 kHz. A custom 3D printed hand fixture prevents wrist flexion and extension & ulnar and radial deviation, serving as the primary mode of interaction for the participant.

### B. Procedure

The experimental protocol consisted of a psychophysical test to determine the JND for a reference linear torsion spring at three different velocities, which varied for each participant. The velocities were based on the participant's preferred exploration velocity ($V_p$) and were set at three levels: **Low** *($V_p$ - 15 deg/s)*, **Natural** *($V_p$)* and **High** *($V_p$ + 15 deg/s)*.

*This material is based upon work supported by the National Science Foundation under NSF Grant# 1657245.

[1]Mohit Singhala is with the Department of Mechanical Engineering, Johns Hopkins University, Baltimore, MD, USA. mohit.singhala@jhu.edu

[2]Jeremy D. Brown is with the Department of Mechanical Engineering, Johns Hopkins University, Baltimore, MD, USA. University. jdelainebrown@jhu.edu

*1) Calibration session:* For the first session, participants explored the virtual torsion spring by pronating (counter-clockwise rotation) then supinating (clockwise rotation) their forearm, avoiding any jerks in their motion, for a total of ten explorations. No restrictions were placed on the magnitude and velocity of exploration. The second to the ninth exploration were used to determine their average maximum angular displacement ($\theta_{max}$) and preferred exploration velocity ($V_p$).

*2) JND Sessions:* Participants performed a simple 2AFC same/different task for three JND sessions, each separated by a five minute break. A weighted 1 up/3 down staircase algorithm was used to determine the JNDs. Based on values suggested by Garcia-Perez [8], the up step-size was set at 10% of the reference stimuli and the ratio of down step-size and up step-size was 0.73 for a proportion correct target of 83.15%. The reference torsion spring was set at 1.5 mNm/deg and the staircase was initialized at a value 2.5 times higher than the reference stimuli. The staircase terminated after ten reversals and the average of the last eight reversals was used to determine the stiffness discrimination threshold. A "1 up/ 1 down" approach was followed until the first reversal. Five catch trials were presented to each participant, where the same reference spring was presented twice. If the participant reported more than one of these trials as "Different", the experiment was terminated.

*3) Exploration Velocity control:* Participants followed the same exploration strategy as in the calibration session, but controlled their velocity using a metronome (for time) and an LED (for displacement). The metronome frequency $f_m$ was set in beats per minute (BPM) according to the desired angular velocity (deg/s) and the maximum angular displacement (deg).

An LED alerted the participant whenever they were within 2.5 degrees of $\theta_{max}$ or the neutral position, to prompt them to either change the direction of rotation or stop. The experimenter alerted the participants to repeat explorations that were not consistent with the required velocity and displacement requirements.

## III. RESULTS

Statistical analysis was limited to nine participants. Two participants were excluded because they failed catch trials, one participant was unable to maintain a smooth motion during calibration, and one participant's session was terminated due to hardware failure. A repeated measuers ANOVA was used to test for the effects of exploration velocity on JND. Wilk-Shapiro and Mauchly's test validated the assumptions of normality and sphericity. We found no statistically significant (F(2,18)=0.349, p>0.05) effect of exploration velocity on JND (see Fig. 2 A)). We also found that participants were very consistent with their preferred angular displacement and velocity for calibration trials (See Fig. 2 B)).

## IV. DISCUSSION

While the sample size is currently limited, our initial results indicate that humans might be able to account for the varying limb dynamics while interacting with the environment. The notion that human interaction with the environment may be guided by the use of an internal model of the motor apparatus by the central nervous system (CNS) has been widely supported in literature [6], [7]. Presently, our results indicate that the same may be true even for our kinesthetic perception of active stimuli, where the force changes based on how we interact with the stimuli. To the best of the authors' knowledge, exploration velocity has not been explored in-depth as an independent variable in haptic perception-based studies and a more thorough investigation is necessary to understand the effects of exploration velocity on kinesthetic perception, or a lack thereof. We are also investigating the possible reasons behind our high JND values. Presently, we believe that the thresholds may be affected by our exploration paradigm (forearm instead of fingers) and the distributed attention required to follow a predefined velocity.

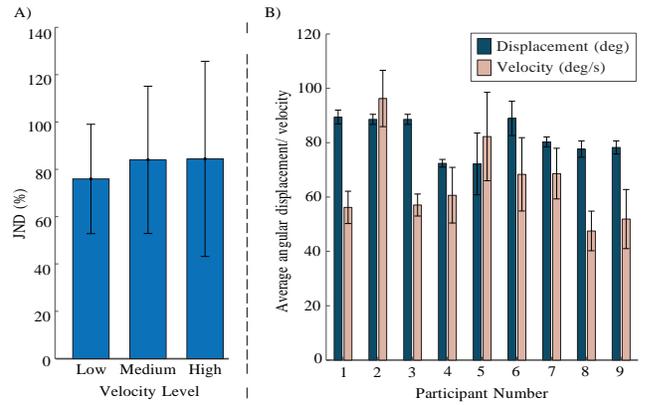

Fig. 2. A) Average JNDs of the nine participants grouped by velocity level: $V_p$ - 15 deg/s (low), $V_p$ (Normal), and $V_p$ + 15 deg/s (High).; B) Average angular displacement and velocity for the calibration trials for each participant.